# A new Integrated Motion Tracking Device (IMTD) for Objective Laparoscopic Training Assessment: Development and Validation

**Siwar Bouzid**
Department of GMSC, Pprime Institute CNRS, ENSMA, University of Poitiers, UPR 3346, France.
Mechanical Laboratory of Sousse (LMS), National Engineering School of Sousse, University of Sousse, Sousse 4000, Tunisia.
siwar.bouzid@univ-poitiers.fr

**Abdelbadia Chaker**
Department of GMSC, Pprime Institute CNRS, ENSMA, University of Poitiers, UPR 3346, France.
Mechanical Laboratory of Sousse (LMS), National Engineering School of Sousse, University of Sousse, Sousse 4000, Tunisia.
Abdelbadia.chaker@univ-poitiers.fr

**Marc Arsicault**
Department of GMSC, Pprime Institute CNRS, ENSMA, University of Poitiers, UPR 3346, France.
Marc.arsicault@univ-poitiers.fr

**Sami Bennour**
Mechanical Laboratory of Sousse (LMS), National Engineering School of Sousse, University
of Sousse, Sousse 4000, Tunisia.
National Engineering School of Monastir, University of Monastir, Monastir 5019, Tunisia.
Computer Engineering, Production and Maintenance Laboratory (LGIPM), University of Lorraine, F-57070, Metz, France.
sami.bennour@eniso.u-sousse.tn

**Med Amine Laribi**
Department of GMSC, Pprime Institute CNRS, ENSMA, University of Poitiers, UPR 3346, France.
med.amine.laribi@univ-poitiers.fr

## ABSTRACT

This paper presents a novel, compact four-degree-of-freedom motion-tracking device (IMTD) designed for training and evaluation in laparoscopic surgery. The device's kinematics, mechanical design, instrumentation, and prototypes are developed and presented to meet the specific requirements of laparoscopic training context, including movement around a fixed center of motion and seamless integration into standard box trainers.

The system IMTD's tracking accuracy and reliability are compared to a motion capture system (MoCap), assessing its ability to capture both angular and translational motions of surgical instruments. The study then focuses on key performance parameters including precision, fluidity, speed, and overall motion efficiency. The results highlight the system's effectiveness in tracking surgical gestures, providing valuable insights into its potential as a tool for training and performance evaluation in minimally invasive surgery. Additionally, IMTD's low cost and integrated design allow for easy integration and implementation in training rooms, offering a practical and accessible solution for general use. By offering objective, real-time feedback, the system can significantly contribute to improving surgical skills and shortening the learning curve for novice students, while also providing a foundation for future development of gesture scoring algorithms and standardized training protocols.

# 1    INTRODUCTION

Laparoscopic surgery requires a specialized set of technical skills that differ significantly from those required in open surgical procedures, thereby demanding dedicated training. The acquisition of minimally invasive surgical techniques is inherently complex, requiring extensive time, dedication, and deliberate practice. However, the operating room constitutes a highly constrained learning environment for surgical residents. The high-stakes setting, time limitations, and the imperative of patient safety restrict opportunities for novices to explore, make errors, and progressively build their competence and confidence. These challenges, widely recognized by surgical educators, have been documented in recent clinical studies, such as Kumar et al. [1], which examined surgeon perceptions of factors affecting the efficiency of conventional and robotic laparoscopy in real-world practice. To address these limitations, surgical simulators have been introduced as educational tools designed to facilitate the development of core laparoscopic skills in a controlled, risk-free environment prior to performing real-life procedures [2], [3].

Different types of simulators have been introduced to address the challenges of traditional surgical training, each tailored to specific learning objectives. Training boxes, for instance, allow residents to practice basic skills such as precision and coordination by monitoring instrument movements. These devices are easy to install, cost-effective and widely available on the market. Often incorporating real surgical instruments, they offer realistic force feedback, enhancing the simulation experience. When equipped with motion-tracking sensors, training boxes offer objective performance evaluation, enabling residents to refine their techniques with measurable feedback. [4], [5] . Virtual reality simulators, on the other hand, offer an interactive environment where surgeons can practice complex procedures and enhance their confidence and technical expertise, particularly through the delivery of detailed tasks metrics and comprehensive feedback on the execution of training exercises. However, their high cost, limited realism, and absence of haptic feedback constrain both their accessibility and training effectiveness [5], [6].

Concurrently, the evaluation of laparoscopic gestures is a crucial component of surgical skill assessment and training. By analyzing the movements of surgical instruments during a procedure, it is possible to quantify the surgeon's performance and identify areas for improvement. Motion analysis techniques significant advancement offered new tool for the objective evaluation of laparoscopic surgical skills. Ebina et al[7]. developed a system that tracks instrument movements during lab simulations, measuring parameters like precision, speed, and errors. Also, Kinematic measurements were adopted for assessing surgical expertise [8]. Research by Oropesa et al. [9] has shown significant correlations between movement parameters and surgical skills providing precise insights into a surgeon's dexterity and coordination.

Kinematic metrics such as time, path length, average speed, and angular path have been identified as essential for objectively assessing surgical performance and helping to differentiate between skill levels. Average speed, for example, is a reliable indicator that reflects the ability of expert surgeons to perform tasks efficiently and accurately, as observed by Hagelsteen et al [10]. However, relying on a single metric is limiting. A more comprehensive evaluation incorporates multiple kinematic measures related to the trajectory, the kinematic of the movement and the simulated surgical task. This multidimensional approach allows for a more accurate and objective assessment of skills, considering aspects such as efficiency, precision, and coordination. Kowalewski et al. [8] demonstrate that combining these measures enhances the objectivity and the reliability of skill assessment and provide a more global assessment of surgical competence [10].

The Peg transfer represents a typical exercise that is commonly used for training. It consists of moving small objects, such as pegs, from one location to another using laparoscopic instruments. Typically, the exercise requires the participant to grasp the peg with one hand, transfer it in the air to the opposite hand, and then place it on a target position on the other side of the board. Target skills are the ability to use both hands independently or collaboratively, which is essential in laparoscopic surgery [11]. Torricelli et al. [12] also points out that bilateral training, which consists of structured exercises designed to improve the coordination of both hands, significantly improves bimanual coordination.

This paper presents a compact, integrated motion tracking device (IMTD) with a remote motion center, developed for laparoscopic surgery training and assessment, as previously introduced [13]. The system provides four degrees of freedom (4DoF), matching typical laparoscopic movements, and is designed to be compatible with standard surgical and training tools.

This paper provides a detailed description of the developed IMTD prototype, including its kinematic configuration, CAD design, instrumentation, and fabrication process. Then its experimental validation against a gold-standard MoCap system (Optitrack), focusing on motion parameter acquisition and comparative metrics. The results section analyzes the recorded data to evaluate the device performance relative to the reference system. This is followed by an application-oriented assessment, where the device is tested in the context of a standard surgical peg transfer task, commonly used for skill evaluation in laparoscopic training. A set of established skill evaluation parameters from the literature is employed to generate a primary assessment of the task related skills, based on the data collected by the IMTD. Finally, results were presented and discussed.

## 2    THE IMTD device kinematics and design

The proposed IMTD device is a standalone system for supporting and monitoring surgical instruments during the learning phases. It is compatible with various existing simulators on the market and can be integrated into both existing and new box trainer structures commonly used in training environments. The device is adaptable to the different standard forms used in these training tools.

Figure 1 illustrates the kinematic model of the proposed structure, developed in previous research [13], which enables the three rotations and one translation required for Minimally Invasive Surgery (MIS) applications. In MIS, the surgeon operates within a constrained, cone-shaped workspace, typically with a 26-degree apex angle [14], demanding precise and controlled instrument movements. The IMTD system is designed to support these movements through the following components: a universal joint providing two rotations ($R_1$, angle $\Phi_1$, and $R_2$, angle $\Phi_2$); a central rotating joint $R_3$, allowing axial rotation $\Phi_3$ of the instrument about the incision point, a fixed center of motion (CoM) where the tool enters the patient's body without lateral displacement; and a prismatic joint (T) enabling translation along the Z-axis. This configuration ensures accurate, constraint-compliant manipulation of surgical instruments.

As mentioned above, a key advantage of our approach is that the tracking system can be seamlessly integrated into the box trainer without requiring any modifications, preserving its full functionality. This non-invasive design makes the solution practical, easy to implement, and compatible with a wide range of training equipment. For our experiments, we used the T5-HD laparoscopic trainer from 3-Med as a validation platform.

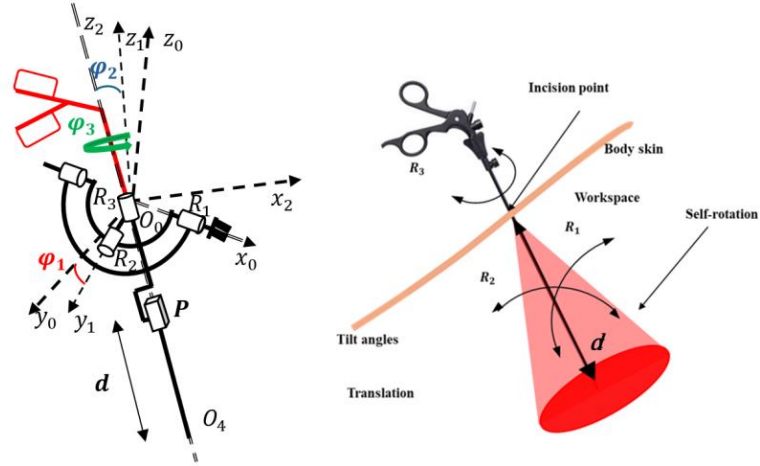

**Fig 1 : Kinematic model of the proposed system with 4 DOF** [13]**.**

The forward kinematic model calculates the surgical tool tip position coordinates as a function of joint variables. It is given by the following equation

$$\mathbf{X} = f(\mathbf{q})$$

(1)

With $\mathbf{X} = [x, y, z]^T$ the vector of Cartesian coordinates and $\boldsymbol{q} = [\phi_1, \phi_2, \phi_3, d]^T$ the vector of joint variables.

Since we can dissociate our system into a gimbal system with translation, we note:

$$[^0T_4] = [^0T_1][^1T_2][^2T_3][^3T_4]$$

(2)

$[^iT_{i+1}]$: represents a homogeneous transformation matrix between two successive "joints":

- $[^0T_1]$: rotation around the X axis by the angle $\phi_1$.

- $[^1T_2]$: rotation around the Y axis by an angle $\phi_2$.

- $[^2T_3]$: rotation around the Z axis by an angle $\phi_3$.

- $[^3T_4]$: translation along Z axis by a distance $d$.

With:

$$[^0T_1] = \begin{bmatrix} 1 & 0 & 0 & 0 \\ 0 & \cos\phi_1 & -\sin\phi_1 & 0 \\ 0 & \sin\phi_1 & \cos\phi_1 & 0 \\ 0 & 0 & 0 & 1 \end{bmatrix} ; [^1T_2] = \begin{bmatrix} \cos\phi_2 & 0 & \sin\phi_2 & 0 \\ 0 & 1 & 0 & 0 \\ -\sin\phi_2 & 0 & \cos\phi_2 & 0 \\ 0 & 0 & 0 & 1 \end{bmatrix} ; \qquad (3)$$

$$[^2T_3] = \begin{bmatrix} \cos\phi_3 & -\sin\phi_3 & 0 & 0 \\ \sin\phi_3 & \cos\phi_3 & 0 & 0 \\ 0 & 0 & 1 & 0 \\ 0 & 0 & 0 & 1 \end{bmatrix} et \ [^3T_4] = \begin{bmatrix} 1 & 0 & 0 & 0 \\ 0 & 1 & 0 & 0 \\ 0 & 0 & 1 & d \\ 0 & 0 & 0 & 1 \end{bmatrix}$$

From the multiplication leading to the homogenous matrix $[^0T_4]$, presented in equation 2, we can directly obtain the final position of the system represented in its fourth column. Hence at a given instant t:

$$\overrightarrow{O_0O_4} = \begin{bmatrix} x(t) = d\sin\phi_2\,(t) \\ y(t) = -d\sin\phi_1(t)\cos\phi_2(t) \\ z(t) = d\cos\phi_1(t)\cos\phi_2\,(t) \end{bmatrix} \qquad (3)$$

### 2.1 IMTD Mechanical design

Figure 2 presents the CAD model of the system with a mounted surgical tool. The system's kinematic architecture is defined by rotational motion around the CoM, precisely aligned with the tool's insertion point. This alignment ensures that instrument movements are directly coupled to the surgical incision, thereby enhancing the accuracy and robustness of the tracking system while complying with the spatial constraint of operating within a 40 mm diameter envelope.

The CoM is achieved through a gimbal mechanism, illustrated in Figure 2b consists of a series of nested, articulated rings that allow controlled rotation about two orthogonal axes. Two 8 mm-diameter magnetic encoders are integrated at the gimbal's pivotal axes to provide accurate feedback on the orientation of the manipulated tool by the practitioner.

The translation movement of the tool is monitored using three segments positioned 120° apart. They maintain uniform contact pressure on the tool shaft using three rollers, supported by an elastic ring, ensuring stable and continuous guidance during tool motion. A magnetic encoder of identical size, mounted on one of the roller shafts, captures the roller's rotational motion, from which the tool's linear translation is accurately derived.

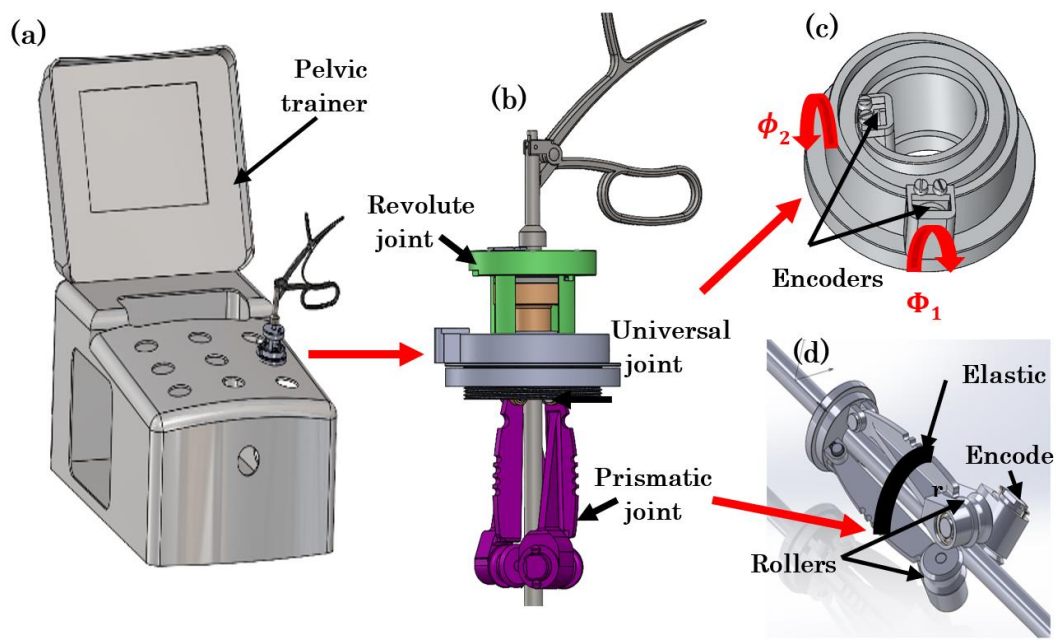

**Fig 2: a) IMTD positioning on a box trainer. b) IMTD Design. c) the gimbal's rotating joints. d) translational guidance system.**

## 2.2   IMTD prototyping and instrumentation

Three high-precision miniature magnetic rotary encoders are integrated to measure the two gimbal rotations and the translation. The collected data is of 10-bit resolution for both angular $\phi_1$ and $\phi_2$ measurements and 9-bit resolution for the translation T value leading to a respective accuracy of 0,35 and 0,7degree, corresponding to an approximate accuracy of 0.055 mm on translations. A fourth rotary encoder positioned at the top of the system providing 12-bit resolution for the self-rotational measurement with an accuracy of 0,08 degree. These accuracy levels closely replicate real laparoscopic procedures, significantly reducing tracking errors and ultimately improving the quality of the collected data.

All four encoders were connected to a custom-designed instrumentation circuit, enabling the acquisition of 4-degree-of-freedom data in real-time using a homemade acquisition and postprocessing interface. Figure 3 details the structure of the electronics.

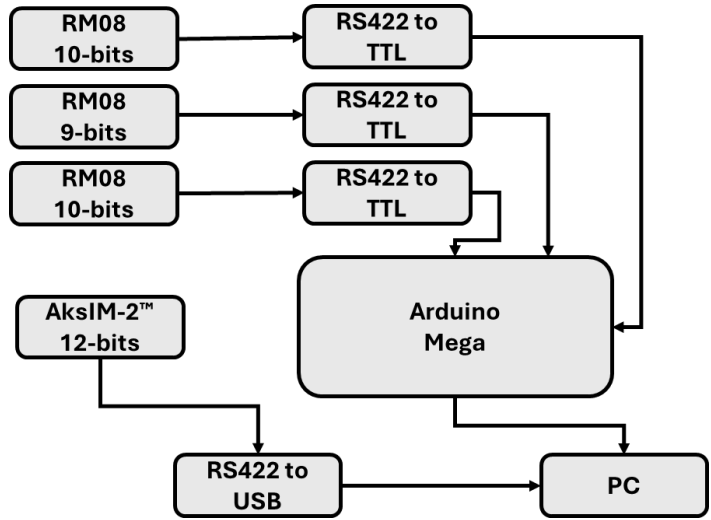

Fig 3: Data acquisition and processing of the IMTD.

Two prototypes of the IMTD systems were developed, focusing on improving resistance and movement. The first one was made with 3D-printed PLA parts, demonstrated the system's kinematics but lacked durability, often breaking under stress or high force applied by surgeons. These improvements resulted in more precise and consistent movement, making the system more reliable and suitable for accurate and repeatable surgical tool evaluation. Figure 4 presents the two versions of the IMTD.

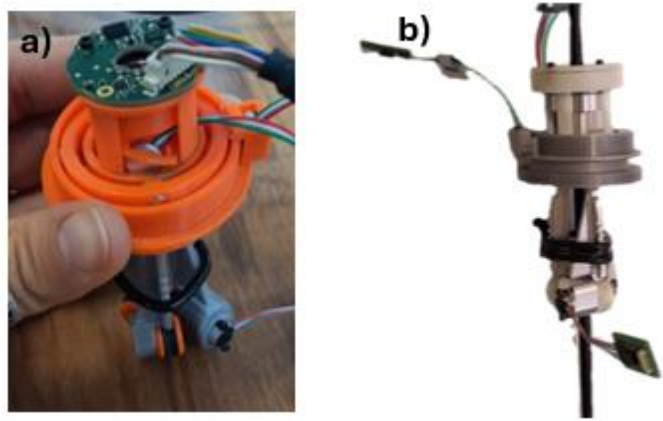

Fig 4: System prototypes: (a) Prototype 1 with PLA components, (b) Prototype 2 with an aluminum guidance system.

## 3    IMTD DEVICE VALIDATION

A comparative analysis of the angular values and translation recorded by the device against those obtained from a gold standard motion capture system (MOCAP) was conducted in order to validate the system performance.

### 3.1    MoCap Data acquisition

A validation setup was established to compare the position and orientation of the surgical instrument tip estimated by the developed system with reference measurements from MoCap system (OptiTrack: V120:Trio). The tool kinematics and movements are determined by tracking a rigid body formed by reflective markers, which are securely affixed to the instrument using a custom 3D-printed holder, and calculates its spatial pose relative to a fixed fiducial frame. Figure 5 presents examples of instrument grippers with attached markers and the corresponding OptiTrack-based data acquisition configuration.

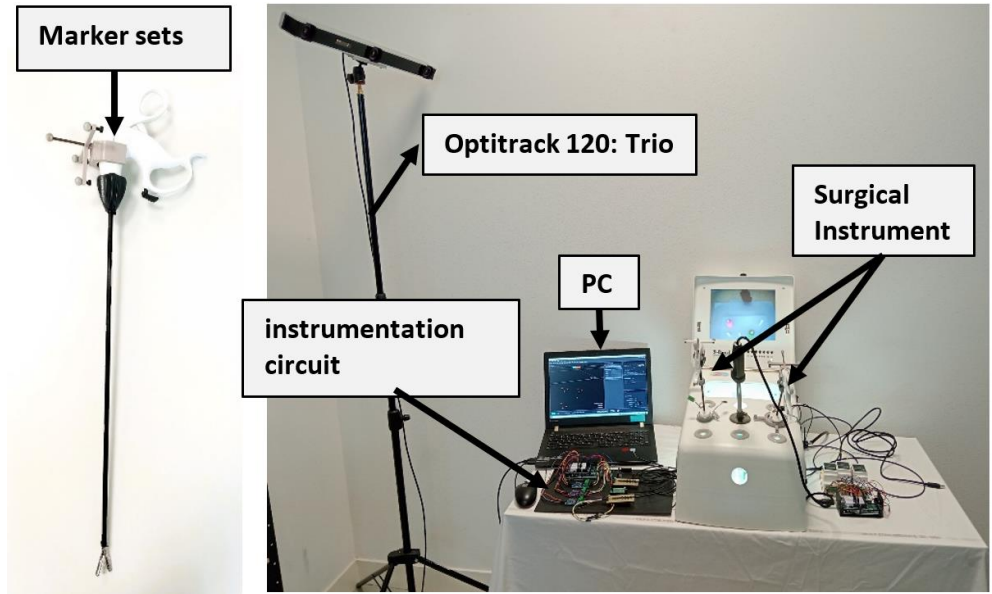

**Fig 5: Acquisition setup of the Surgical instrument with markers attached, tracked by the OptiTrack V120: Trio system.**

Figure 6 shows the experimental setup used to align the reference frames of our system and the OptiTrack system. Markers on the box trainer (Figure 6-a) and the surgical tool (Figure 6-b) defined their respective reference frames, with an additional marker on the tool tip for static distance measurements.

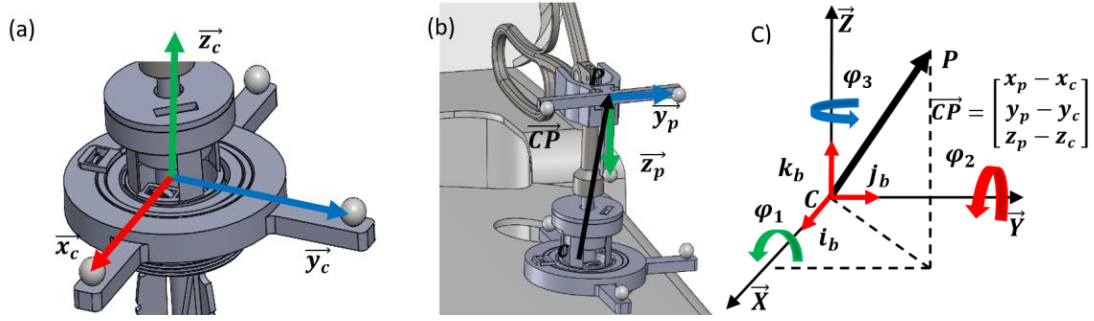

**Fig 6: (a) Surgical tool with markers for the center of reference. (b) Device with markers defining the center of reference for the motion capture system. (c) Schematic of Base Reference Frame and presentation of vector V.**

The frames in the setup are as follows:

- **System center:** $C = (x_c, y_c, z_c)$ the origin of the reference frame in the global coordinate system.
- **Base frame**: Defined by three orthogonal unit vectors $i_b, j_b, k_b$.
- **Tool point**: $P = (x_p, y_p, z_p)$ a known position in the global frame.
- **Vector V**: Defined $\mathbf{V} = \overrightarrow{CP}$, calculated as $(x_p - x_c, y_p - y_c, z_p - z_c)$.

The IMTD system kinematics can be represented using a homogeneous transformation matrix T:

$$T = \begin{bmatrix} R & t \\ 000 & 1 \end{bmatrix} \tag{5}$$

Where:

$\boldsymbol{R}$ is a $3 \times 3$ rotation matrix. It is defined as: $\boldsymbol{R} = R_z(\phi_3) R_y(\phi_2) R_x(\phi_1)$ This corresponds to successive rotations: first about the x-axis by $\phi_1$, then the y-axis by $\phi_2$, and finally the z-axis by $\phi_3$.

$\boldsymbol{t}$ is the translation vector $[t_x, t_y, t_z]$, representing the position of the object with respect to the reference frame.

The last row $[000 \quad 1]$ makes the matrix homogeneous $(4 \times 4)$, allowing rotations and translations to be combined into one matrix operation.

The translation vector $V$ can be directly calculated from $\overrightarrow{CP}$ as it represents the relative position between the center C and the point P:

$$\mathbf{t} = \begin{bmatrix} x_{pc\_initial} - x_{pc\_final} \\ y_{pc\_initial} - y_{pc\_final} \\ z_{pc\_initial} - z_{pc\_final} \end{bmatrix} \tag{6}$$

Using the vector $V$ and the base frame axes ($\boldsymbol{i_b}, \boldsymbol{j_b}, \boldsymbol{k_b}$) the angles can be computed as follows:

$$\phi_1 = \mathrm{atan2}\big(v_y, v_z\big), \ \phi_2 = \mathrm{atan2}\big(v_x, v_z\big), \ \phi_3 = \mathrm{atan2}\big(v_y, v_x\big) \tag{7}$$

During the validation tests, a static acquisition was first performed to determine the system's mechanical zero angles and compute the transformation matrix between the system and camera frames. This was followed by identical acquisition protocols for both the left- and right-hand systems, involving a full workspace scan by varying gimbal angles $\phi_1$ and $\phi_2$, translation $t$ and self-rotation $\phi_3$.

In the IMTD Systems, gimbal angles are computed directly, whereas in the motion capture system, they are derived from the motion vector $\mathrm{V} = \overrightarrow{CP}$ between the system center and the instrument tip.

### 3.2   Results

Figure 7 shows the 3D reconstruction of the left tool trajectory in the workspace using the developed systems. A mentioned above, the performed test maps the complete workspace or "cone of operation" in order validate the performance on all points.

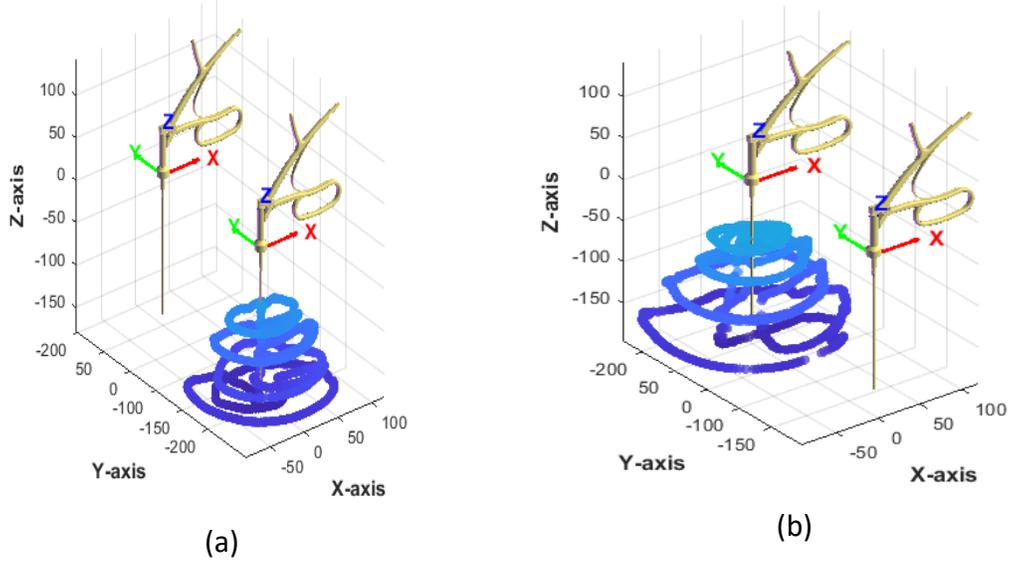

**Fig 7: 3D Tool trajectory reconstruction using the IMTD systems (a) Left tool (b) Right tool.**

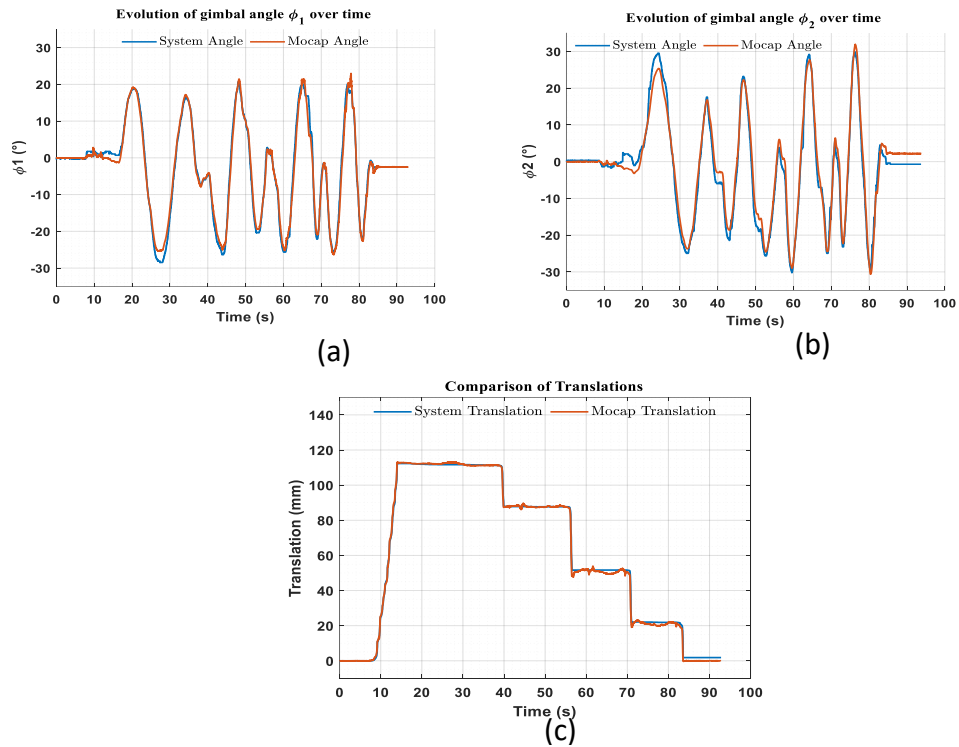

**Fig 8: Comparison of the system-generated data with motion capture (MoCap) measurements for the left hand. (a) Gimbal angle $\phi_1$, over time, (b) Gimbal angle $\phi_2$, (c) Translation T in millimeters.**

The three graphs, shown in figure 8, provide a comparison between the system-generated data and motion capture (MoCap) measurements, evaluating the system's performance in replicating angular and translational motions. While minor deviations appear during rapid oscillations, the overall alignment remains acceptable. The Mean Squared Error (MSE) for $\phi_1$ is 1,93 $degrees^2$, showcasing the system's ability to effectively capture angular dynamics. The MSE for $\phi_2$ is 4,8 $degrees^2$, reflecting higher differences in against MoCap data. These discrepancies. Figure 8-c examines translations $T$ in millimeters, where the system achieves an almost perfect match with MoCap data, even during abrupt transitions. The MSE for $T$ is 1 $mm^2$, underscoring the system's performance in capturing translational motion with high fidelity. In summary, the system accurately reproduces both angular and translational motions, with minor limitations during rapid angular changes validating its reliability and suitability for the surgical training applications.

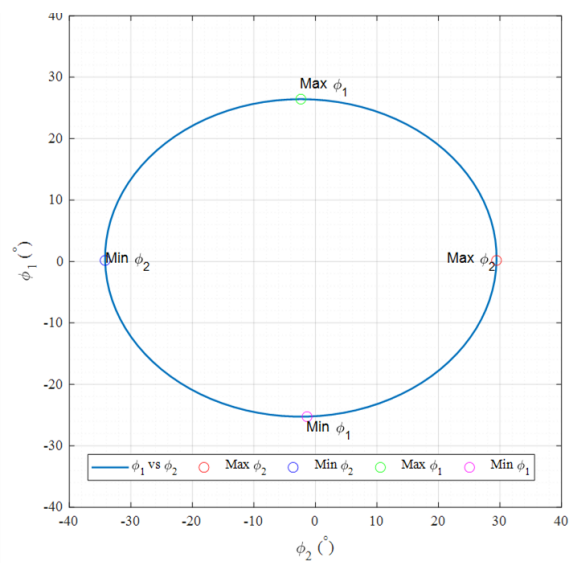

**Fig 9: Workspace cone base formed by $\phi_1$ and $\phi_2$ variation.**

Figure 9 illustrates the boundary of the workspace prescribed by the instrument using the IMTD system. The relationship between the two Cardan angles, $\phi_1$ and $\phi_2$, for the left-hand system forms an elliptical curve, effectively representing the system's maximum range of motion. Notably, this workspace remains within the 26-degree conical constraint typical of laparoscopic procedures, ensuring that the motion stays within the predefined surgical tasks limits.

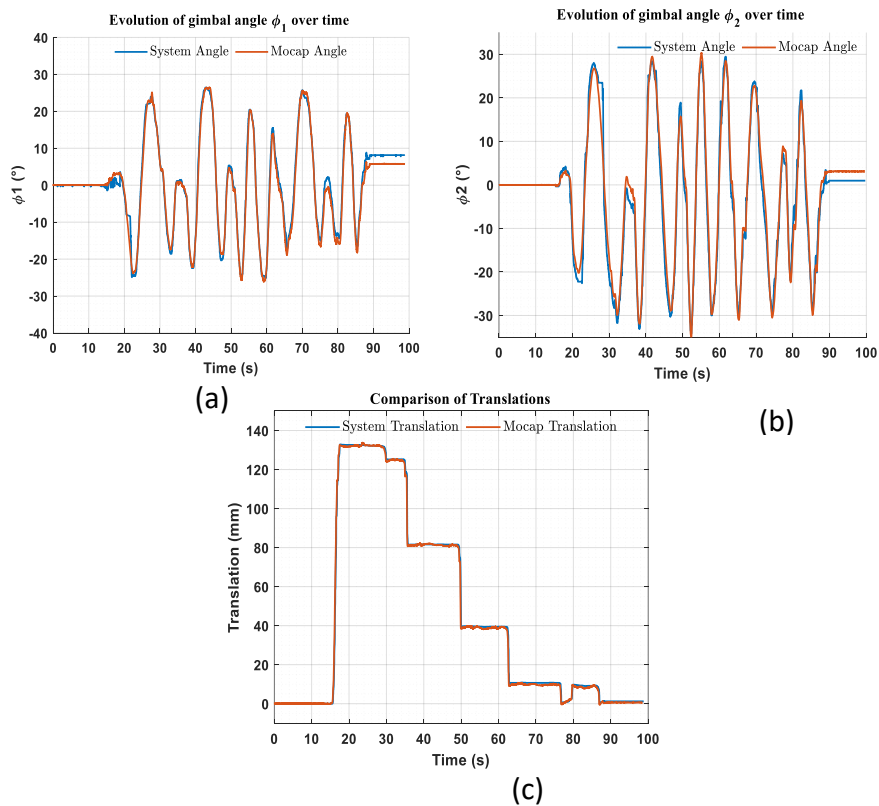

(a)

(b)

(c)

**Fig 10: Comparison of the system-generated data with motion capture (MoCap) measurements for the right hand. (a) Gimbal angle $\phi_1$ over time, (b) Gimbal angle $\phi_2$. (c) Translation T in millimeters.**

Figure 10 represents the evolution of angular and translational data for the right IMTD System, presenting similar performance to the left one. The Mean Squared Error (MSE) for $\phi_1$ is 2,83 $degrees^2$, reflecting the system's precision. Both the system and MoCap follow a consistent pattern, though minor phase and amplitude mismatches are noticeable during sharp changes. The MSE for $\phi_2$ is approximately 2 $degrees^2$ .For the translations, both data exhibit an almost identical trend, even during rapid transitions, the system delivering good accuracy in reproducing translational motion. The MSE for translations $T$ is 2,68 $mm^2$ .Overall, the results validate the second system's capability to replicate angular and translational motions, with minor limitations observed during rapid angular variations.

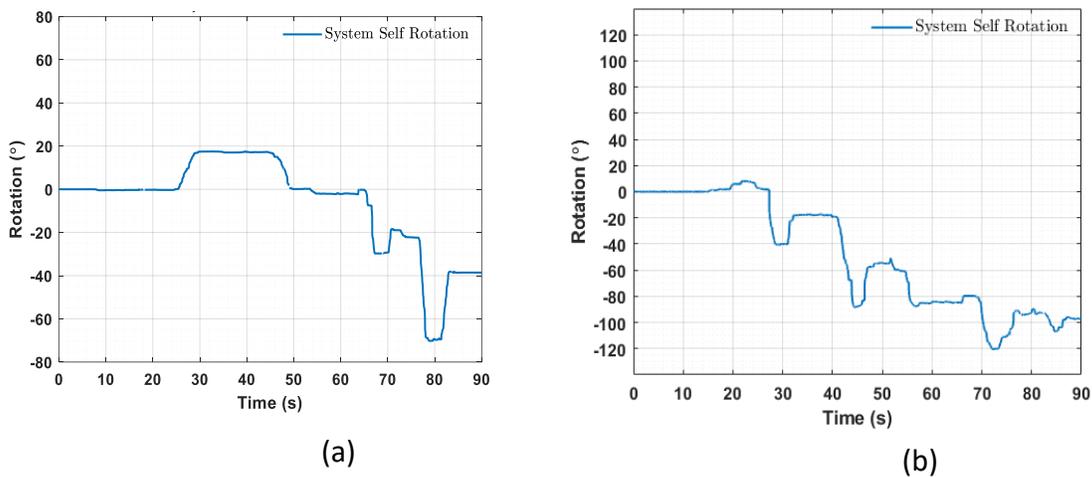

**Fig 11: Evolution of System Self-Rotation Over Time $\phi_3$ :(a) for the left hand, (b) for the right hand.**

Figure 11 shows the evolution of the $\phi_3$ system's self-rotation over time, represented in degrees, for both the left and right hand. Unlike the other motion parameters, no comparison with Optitrack data was made for $\phi_3$, as it is decoupled from the other degrees of freedom and the magnetic rotation sensor used offers superior accuracy than the Optitrack system. Therefore, the graphs focus solely on the self-rotation values generated by the system itself. The left-hand system, shown in figure 11-a, exhibits self-rotation $\phi_3$ values ranging from +40° to -70°, with notable transitions and steady states observed throughout the period. These dynamic changes provide valuable insight into how the system responds to internal factors, offering a unique view of its behavior. Meanwhile the right-hand system, shown in Figure 11-b, demonstrates significant variations, ranging from +10° to -120°. Although both hands show transitions between stable states, the right-hand system shows more pronounced oscillations and a larger negative range.

These oscillations in self-rotation are likely caused by the validation test itself, as the graphs are determined directly from the motion data, reflecting natural adjustments made during the testing process.

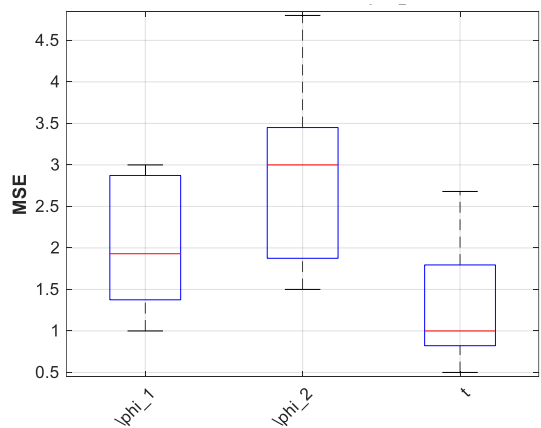

**Fig 12: Distribution of Mean Squared Errors (MSE) for the Parameters $\phi_1$, $\phi_2$, and Translations $t$.**

The box plot, shown in figure 12, compares the mean square error (MSE) for three parameters: $\phi_1$, $\phi_2$, and $t$. A total of 15 tests were included in this analysis, following the same approach as the trajectory comparison with the MoCap system. Among the three parameters, $\phi_2$ has the highest median RMSE (approx. 3 $degrees^2$) and the greatest variability, indicating significant tracking errors. $\phi_1$, has a moderate median RMSE (approx. 2 $degrees^2$) with less variation, while $T$ has the lowest median RMSE (approx. 1 $mm^2$) and is the most stable. Overall, although $\phi_2$ demonstrates higher errors, these discrepancies remain within acceptable limits and do not substantially affect the calculation of motion parameters. The observed variations are likely due to differences in how the MoCap system computes marker positions compared to the IMTD. Potential issues related to sensor positioning and mechanical clearance have been minimized thanks to the sensors resolution and the machining of all functional parts. Residual discrepancies may still arise from algorithmic processing delays, but these have a limited impact on overall motion measurement accuracy in the current use case.

## 4      EXPERIMENTAL VALIDATION ON SURGICAL TRAINING EXERCISE

This section describes the methodology for evaluating laparoscopic gestures using the data collected from the IMTD systems, focusing on the movements of two equipped hands and their related tool trajectory metrics during a peg transfer exercise.

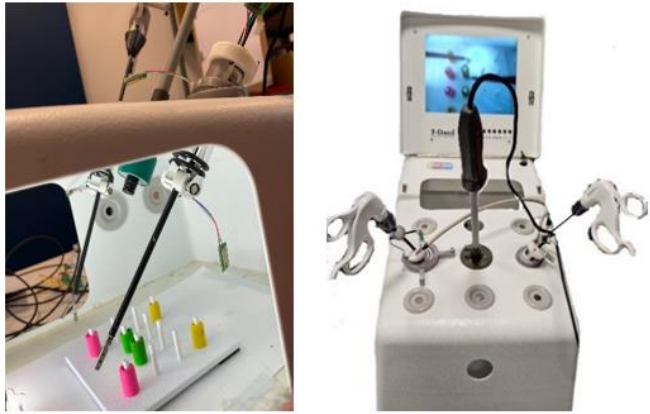

**Fig 13: Configuration of Dual Motion-Tracking Systems Mounted on the Pelvic trainer.**

The peg transfer exercise, illustrated in figure 14, requires the trainees to move six pegs from a group on their right side to a set of poles on their left side in one direction using laparoscopic instruments. The task must be completed in a single direction, meaning that each peg is transferred sequentially, first by grasping it with one hand as, passing it to the opposite hand in mid-air, and then precisely placing it onto the corresponding pole. Although the task appears simple, it is designed specifically to develop essential skills for surgery students such as precision, hand-eye coordination and dexterity. For our system, this exercise served as an initial test to determine the motion tracking parameters, which were selected based on parameters identified in previous research[13]using this exercise.

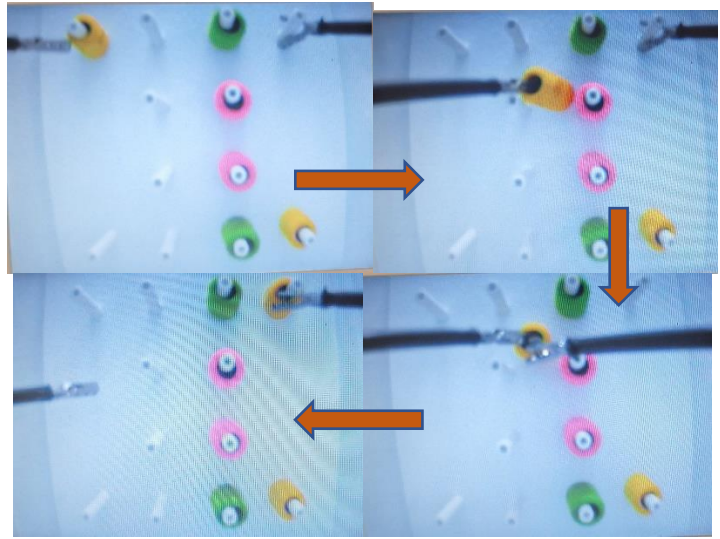

**Fig 14: Representation of the peg transfer exercise.**

### 4.1 Surgical gesture evaluation parameters

The four DoF data of the test movements were collected using the dual setup of the IMTD system shown in Figure 13. Instrument movements are then determined by applying forward kinematic models to the tracking data from our four sensors, as presented in the previous sections. These models, combined with sensor angular measurements and system parameters, allow us to calculate the movements of the tools and to extract multiple metrics that will be used to objectively characterize and evaluate the trainee performance:

- Time :The total time taken to complete the task (s):

$$Time = t_{final} - t_{initial} \qquad (8)$$

- $IT$ (%) - Idle Time Percentage:

$$IT = \frac{T_{idl}}{Time} \times 100 \qquad (9)$$

$T_{idl}$: Sum of the periods when the system is stationary or paused.

- Path length ($L$) : Total Distance traveled by the end of the instrument (m):

$$L = \sum_{i=1}^{n-1} \sqrt{(x_{i+1} - x_i)^2 + (y_{i+1} - y_i)^2 + (z_{i+1} - z_i)^2} \qquad (10)$$

- The Depth Workspace (DW), representing the range of depth in which the tool traveled during the task, can be expressed as:

$$DW = h = z_{max} - z_{min} \qquad (11)$$

$z_{max}$: is the maximum depth reached by the instrument along its axis,

$z_{min}$: is the minimum depth reached by the instrument along its axis.

- Average speed $AS$ ($mm/s$): The average speed of movements, typically measured as the ratio of distance:

$$AS = \frac{L}{Time} \qquad (12)$$

- Average acceleration $AA$ ($mm^2/s^2$): Rate of change of instrument speed within the setting

$$AA = \frac{\sum_{i=1}^{n-1} |v_i - v_i|}{Time} \qquad (13)$$

- Fluidity of movement (MF): refers to the smooth, continuous, movement of the tool without abrupt stops or trembling related vibration. It is often linked to jerk, which is the rate of change of acceleration.

$$MF = \frac{1}{Jerk} \qquad (14)$$

$$Jerk = \frac{1}{T} \int_{t=0}^{T} \frac{d^3 |r(t)|}{dt^3} dt \qquad (15)$$

The Jerk quantifies the variation in acceleration over time. A low or zero jerk indicates smoother, more fluid motion, while higher jerk values reflect abrupt changes in acceleration, which can result in jarring or unsteady movement.

- Volume $V$ ($mm^3$). It is defined as the area enclosed by the farthest positions reached by the instrument throughout the task.

$$V = \frac{1}{3}\pi h(R^2 + Rr + r^2) \tag{16}$$

$h$: distance from the trocar to the deepest point reached by the instrument.

$R$; maximum reachable radius at the deepest point.

$r$: radius at the trocar level .

Building on this framework, we have classified the motion tracking parameters into three subcategories to ensure a more comprehensive evaluation of surgical performance as presented in the table 1. The first subcategory, **Execution Rapidity**, focuses on temporal efficiency and includes parameters such as total task duration and inactivity time, reflecting the rhythm and flow of the procedure. The second subcategory, **Gesture Control**, assesses the surgeon's precision and control through parameters such as speed, acceleration, and movement fluidity, which highlight the smoothness and coherence of movements. Finally, **3D Navigation in the Working Area** evaluates spatial awareness and efficiency using parameters such as distance traveled, depth of motion, and the volume of workspace explored, measuring the surgeon's ability to navigate effectively within the constraints of the operating area. This classification system provides valuable insights for training, evaluation, and optimization of laparoscopic techniques.

These metrics and sub-families were then used to develop evaluation measures for surgical gestures, implemented with our motion tracking system incorporating both sensor-equipped hands. This system allows in-depth analysis of bimanual coordination and precision, providing an innovative tool for training and evaluation in the field of laparoscopic surgery.

**Table 1: Motion tracking parameters subcategories**

| Subcategories | | |
|---|---|---|
| Execution Rapidity | Gesture Control | 3D Navigation in the Working Area |
| Total task duration | Acceleration | Distance traveled |
| Inactivity time | Movement fluidity | Depth of motion |
| | | Volume of the workspace |

### 4.2  Experimental results

In this section, we describe the set-up of a real peg transfer exercise performed using the IMTD systems. The main objective of this exercise is to collect motion data to determine the key kinematic parameters used in the evaluation of laparoscopic gestures.

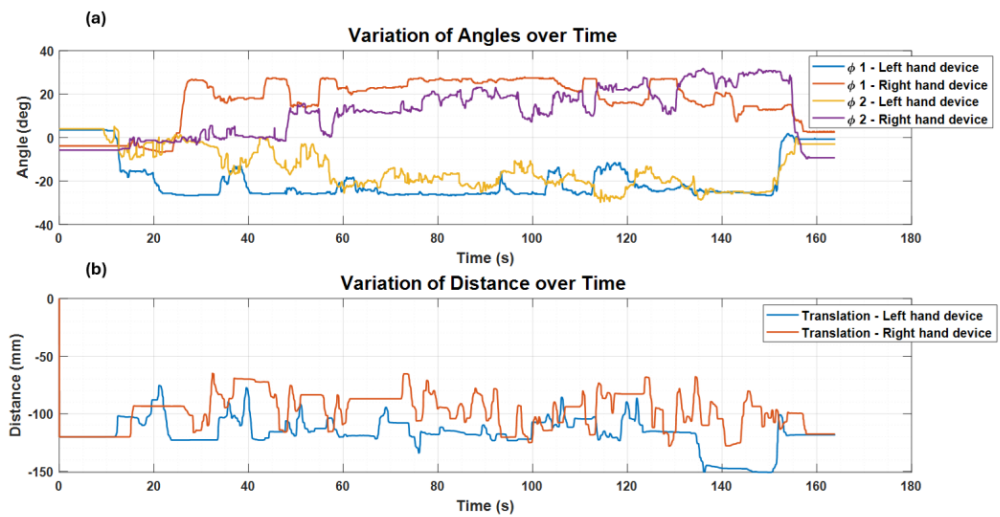

**Fig 15: Angular and Translational Variations for Left hand device and right-hand device.**

Figure 15 shows the variations of angles ($\phi_1$ and $\phi_2$) and translation distances for two IMTD systems (Right Hand Device and Left Hand Device) over time. The angular variations ($\phi_1$ and $\phi_2$) show the distinct behavior of each hand. The left-hand device shows oscillations mainly in the negative range, indicating its role in stabilizing movements, while the right-hand device shows greater fluctuations, particularly between 60s and 100s, suggesting a critical phase of the task requiring precise adjustments. After 160 s, the angles stabilize, marking the end of the movements sequence. Regarding the translational movements, the right device shows more pronounced oscillations, indicating its dominant role in object manipulation, while the left device provides support and coordination. The stabilization of both curves after 160 seconds suggests the end of the task.

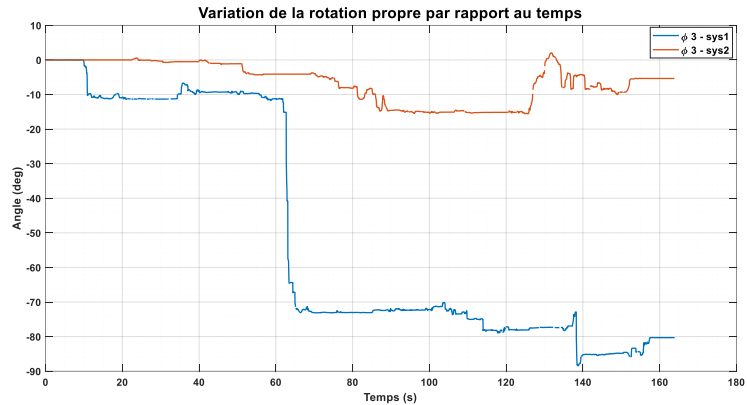

**Fig 16: Comparative Rotational Behavior of Right Hand Device and Left Hand Device During Peg Transfer Exercise: Right vs Left Hand Dynamics.**

Figure 16, highlights distinct rotational behaviors between the two systems during the realized tests. Right Hand Device ($\phi_3$−Right Hand Device) exhibits a significant and abrupt rotational change around $t = 50$, transitioning to stabilization at a lower angular value (-70° to -80°). This important variation probably indicates realignment or repositioning of the surgical instrument in the user's right hand. In contrast, Left Hand

Device ($\phi_3$–Left Hand Device) shows a smoother and more gradual progression, maintaining a consistent angular trend with minor oscillations emerging after around $t = 120$. This steadier behavior suggests that the left hand is performing less abrupt or compensatory movements, serving a more stabilizing or supportive role during the surgical procedure.

The data acquired, along with the kinematic modeling presented in the first section, were then used to compute the forward kinematic model of the IMTD systems to determine the position of the tool tip for both hands present in figure 17-a and 17-b. The resulting data was then applied to calculate the motion tracking parameters and to characterize the laparoscopic gesture families developed previously for evaluation.

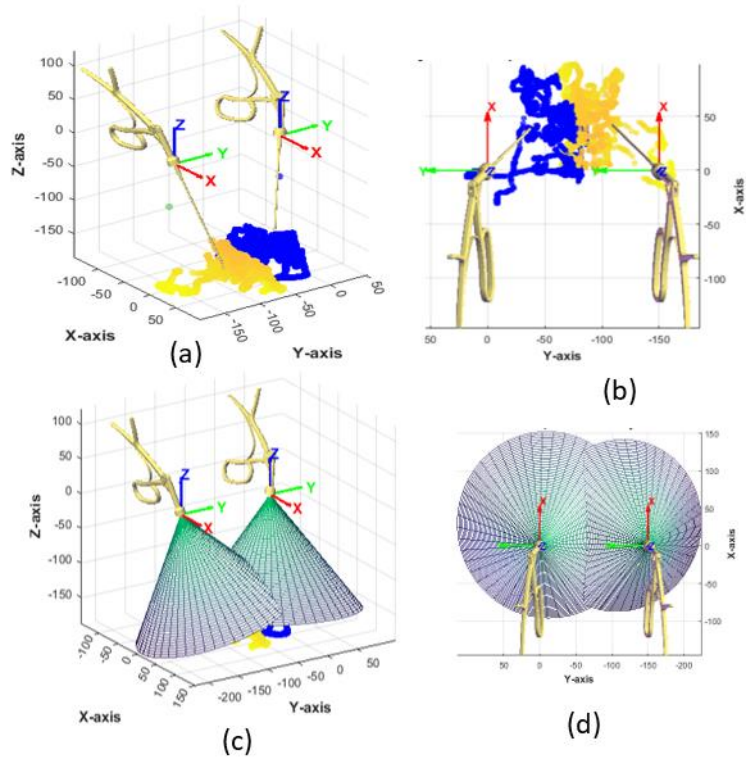

**Fig 17: a) and b): Kinematic results for tool tip positioning in the 3R1T system. c) and d): Cones of the left and right instrument workspaces.**

Figures 17-c and 17-d show the workspace cones of the two instruments, illustrating key parameters for assessing the operational field and tool accessibility. In particular, Figure 17-d reveals that the workspace of the right hand is smaller than that of the left hand, a difference probably due to the dominance of the user's right hand. This disparity highlights the effect of hand dominance on spatial reach and handling within the system, which influences the calculation of various performance parameters and scores which will be determined after.

**Table 2: Comparison of Performance Metrics for Dominant (Right) and Non-Dominant (Left) Hands in Laparoscopic Tasks.**

| Metrics | Left Hand Device | Right Hand Device |
|---|---|---|
| $T(s)$ | 164 | 164 |
| $IT(\%)$ | 52 | 49 |
| $L(mm)$ | 2043 | 1857 |
| $DW(mm)$ | 55 | 56 |
| $AS(mm/s)$ | 12,47 | 11,33 |
| $AA(mm^2/s)$ | 0,076 | 0,069 |
| $Jerk$ | 163,73 | 155,66 |
| $FM$ | $61*10^{-4}$ | $64.2*10^{-4}$ |
| $V(mm^3)$ | 463591 | 424029 |

Table2 presents a detailed comparison of the parameters evaluated for the right and left hands, providing a comprehensive analysis of their performance in laparoscopic tasks. The system labelled Right Hand Device corresponds to the dominant right hand, which is often the most dexterous hand. In contrast, the performance of the left hand, assessed using the same parameters, represents the non-dominant hand Left Hand Device.

The comparison of performance between Left Hand and Right Hand IMTD Devices highlights key differences in efficiency and motion dynamics, which can be analyzed through the subcategories of Execution Rapidity, Gesture Control, and 3D Navigation in the Working Area. In terms of Execution Rapidity, both systems show similar task completion times (164 seconds for both), but Left Hand Device has a slightly higher idle time percentage (52% vs. 49%), indicating longer periods of inactivity. Regarding Gesture Control, Left Hand Device covers a greater total distance (2043 mm vs. 1857 mm for Right Hand Device), indicating a less optimized trajectory, even though it achieves a higher average speed (12,47 mm/s vs. 11,33 mm/s for Right Hand Device), suggesting that Left Hand Device prioritizes speed over precision. Finally, in terms of 3D Navigation in the Working Area, Left Hand Device operates within a larger motion volume (463591 mm³ vs. 42402 mm³ for Right Hand Device), emphasizing less spatially efficient movements. The results showed a mean Jerk of 164 for the Left Hand and 156 for the Right Hand, indicating comparable smoothness (FM) between both hands. Furthermore, the average acceleration (AA) metrics were nearly identical for both systems, reflecting similar motion dynamics. In summary, while Left Hand Device favors speed (with faster movements), Right Hand Device demonstrates better spatial efficiency and a more optimized trajectory, highlighting a trade-off between speed, precision, and trajectory optimization in the two systems.

Using the parameters identified, we can systematically assess and compare workspace efficiency and the performance of both hands during laparoscopic procedures.

This comparison provides a better understanding of bimanual coordination, an essential skill in minimally invasive surgery, where both hands must work in complementarity to achieve precision and efficiency.

## 5    CONCLUSION

In conclusion, the tool movements in both systems demonstrated strong kinematic tracking capabilities, with precise angular and translational behavior over time. Each hand's system smoothly combined rotational and translational motions, essential for replicating complex surgical gestures.

The angular motion, characterized by the evolution of the gimbal angles $\phi_1$ and $\phi_2$ highlights the tool's ability to perform smooth rotations and dynamically adapt to changes in orientation. The high degree of correspondence with motion capture (MoCap) data underscores the system's tracking accuracy. Minor discrepancies observed during rapid oscillations suggest robust performance under dynamic conditions, with slight deviations only during high-frequency variations.

Similarly, the translational motion demonstrates exceptional precision, maintaining near-perfect alignment with MoCap data, even during abrupt position changes.

Our results confirmed the feasibility, validity and reliability of the proposed system as a practical tool for training and performance evaluation. These transitions were achieved with minimal error, further emphasizing the tool's adaptability and reliability.

Future advances in this technology could include extended validation to a variety of surgical tasks and the development of standardized assessment measures to improve the quality of surgical training and skill acquisition. In addition, the system offers the possibility of creating a comprehensive grading application, exploiting data from expert surgeons as benchmarks to provide accurate and objective feedback to trainees, thus optimizing the learning process and enhancing skill refinement.

Beyond its accuracy, the system stands out for its cost-effectiveness and ease of integration. Designed to be mounted on a pelvic without modifications, it can be easily duplicated in training environments. Its simplicity allows for quick setup and removal, making it an accessible tool for self-learning.

Based on this framework, we are developing a scoring algorithm that compares trainees' kinematic data with expert references, assessing accuracy, speed and fluidity. This system will identify areas for improvement and classify performance, simplifying skill development. Further validation and standardized measurements will refine its applications, while real-time feedback could accelerate learning and improve surgical proficiency.

Additionally, our system incorporated advanced 3D navigation capabilities that assess trainees' spatial navigation, hand-eye coordination, and fine motor control. By optimizing movement efficiency and minimizing unnecessary actions, it provides targeted feedback to refine essential laparoscopic skills.